\documentclass[conference]{IEEEtran}
\IEEEoverridecommandlockouts
\usepackage{cite}
\usepackage{amsmath,amssymb,amsfonts}
\usepackage{algorithmic}
\usepackage{graphicx}
\usepackage{textcomp}
\usepackage{xcolor}
\usepackage{booktabs}
\usepackage{multirow}
\usepackage{amssymb}

\usepackage{booktabs}
\usepackage{graphicx}

\usepackage[table]{xcolor}
\definecolor{pastelgreen}{HTML}{E6F4EA}

\def\BibTeX{{\rm B\kern-.05em{\sc i\kern-.025em b}\kern-.08em
    T\kern-.1667em\lower.7ex\hbox{E}\kern-.125emX}}
\begin{document}

\title{Boosting Neural Video Codec via Scale-Driven Online Flow Refinement}

\author{
\textit{Tiange Zhang}$^{1,2,4,\ast}$\thanks{$^{\ast}$\scriptsize Equal contribution. $^{\dagger}$Corresponding authors: Xiandong Meng and Zhimeng Huang. This work was supported in part by R24115SG MIGU-PKU META VISION TECHNOLOGY INNOVATION LAB, the Key Research \& Development Program of Peng Cheng Laboratory under grant PCL2024A02, Beijing Natural Science Foundation No. 4264094 and New Cornerstone Science Foundation through the XPLORER PRIZE.}, 
\textit{Rongqun Lin}$^{3,\ast}$, 
\textit{Haocheng Tang}$^{4}$, 
\textit{Xiandong Meng}$^{2,\dagger}$,\\ 
\textit{Weijia Jiang}$^{5}$, 
\textit{Zhimeng Huang}$^{4,\dagger}$ and 
\textit{Siwei Ma}$^{1,2,4}$\\
$^{1}$Peking University Shenzhen Graduate School \quad
$^{2}$Pengcheng Laboratory \quad
$^{3}$City University of Hong Kong\\ 
$^{4}$School of Computer Science, Peking University \quad
$^{5}$Migu Interactive Entertainment Co., Ltd\\
\texttt{\{tgzhang, hctang\}@stu.pku.edu.cn, rqlin3-c@my.cityu.edu.hk, mengxd@pcl.ac.cn}\\ 
\texttt{jiangweijia@migu.chinamobile.com, \{zmhuang, swma\}@pku.edu.cn}
}
\maketitle

\begin{abstract}
Although state-of-the-art neural video codecs (NVCs) have achieved remarkable performance, they suffer from limited generalization when encountering complex motion patterns unseen during training. To bridge this domain gap without the expensive cost of online fine-tuning, we propose a Training-Free Scale-Driven Online Flow Refinement (SOFR) method. Serving as a plug-and-play module, SOFR integrates motion information from coarse and fine scales and dynamically fuses them according to warping accuracy, effectively rectifying motion estimation errors with negligible computational overhead. Furthermore, we design a rate-aware strategy that selects different dynamic fusion strategies according to bitrate modes, and employs a reliability check based on warping error to ensure robustness. Extensive experiments on the USTC-TD dataset verify the effectiveness and generalization of SOFR across various NVC frameworks, including DCVC-SDD, DCVC-FM, and EHVC. Notably, it brings an average of 2.84\% and 4.05\% bitrate savings in terms of PSNR and MS-SSIM, respectively, to DCVC-FM with negligible coding time increase. Our code is available at https://github.com/SunnyMass/SOFR.
\end{abstract}

\section{Introduction}
\label{sec:intro}

In recent years, the explosive growth of high-definition video traffic has placed immense pressure on internet bandwidth, necessitating more efficient video compression technologies. While traditional standards like H.266/VVC~\cite{bross2021overview} have largely plateaued in coding efficiency, Neural Video Codecs (NVCs)~\cite{ma2019image,jia2025emerging,LIN2023126396} have emerged as a promising alternative. By leveraging pure neural network architectures in an end-to-end manner, state-of-the-art NVCs have demonstrated rate-distortion (R-D) performance comparable to or even surpassing traditional codecs.

However, the practical deployment of NVCs faces a critical challenge: the domain gap between training datasets and real-world inference scenarios. Most NVCs are trained on static datasets with limited motion diversity (e.g., Vimeo-90k~\cite{xue2019video}). When exposed to complex, unseen motion patterns during inference, the pre-trained optical flow networks often fail to generate accurate motion vectors (MVs). Inaccurate motion estimation results in imprecise warped contextual features. This misalignment disrupts effective temporal feature propagation, thereby limiting coding gains.

To mitigate this domain shift, recent research has explored various adaptation strategies. Some works~\cite{cong2025integrating,tang2024offline,agustsson2020scale,hu2022coarse} focus on \textit{offline optimization}, incorporating advanced optical flow constraints or motion priors during the training phase to enhance generalization. However, these models remain fixed and lack the flexibility to adapt to the specific motion statistics of unseen test videos.

Other approaches employ \textit{online optimization} \cite{tang2024offline, chen2024group, oh2024parameter,11561917} to achieve instance-level adaptation, where network parameters are updated via back-propagation to minimize the R-D cost. While effective, this process brings significant computational overhead and latency. Alternatively, training-free methods~\cite{zhai2025lbvc,bilican2025content,yilmaz2024motion} attempt to optimize motion representation by exhaustively searching through multiple down-sampling scales. 
However, these methods typically operate on a rigid selection mechanism, choosing only the single best scale while discarding others. This overlooks the inherent complementarity between granularities: coarse-grained flows ensure robust structural stability, while fine-grained flows preserve intricate local details. By failing to synergize these conflicting yet complementary signals, existing approaches cannot achieve the optimal balance required for precise motion alignment.

\begin{figure*}[ht!]
  \centering
  \includegraphics[width=0.98\linewidth]{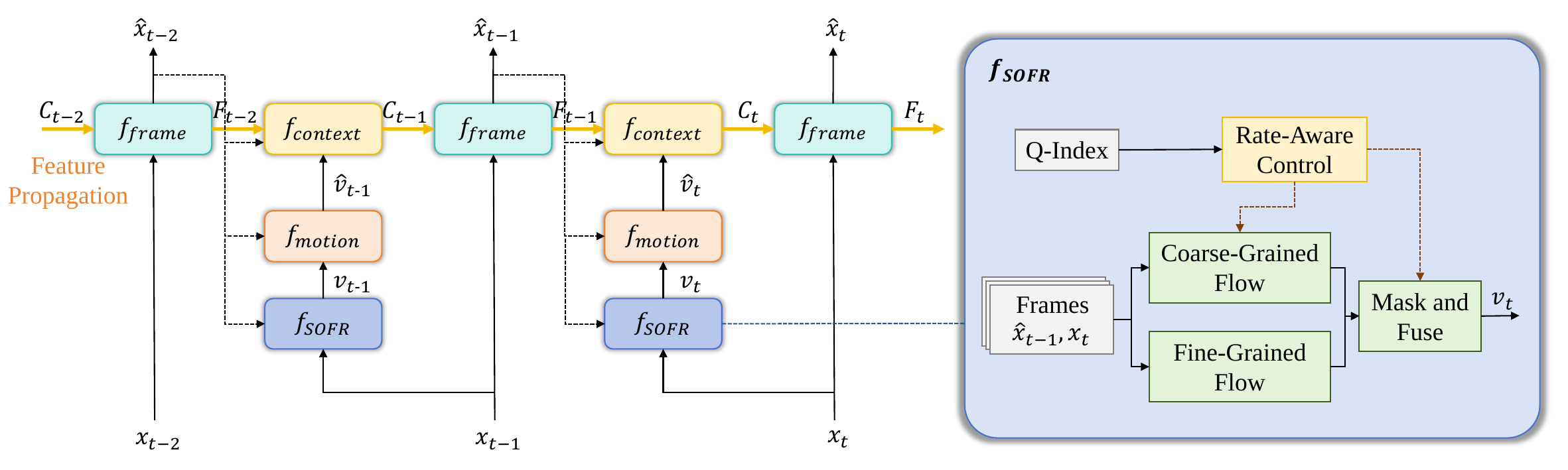}
  \caption{Overview of the proposed neural video compression framework. The left panel depicts the temporal context propagation pipeline. At each timestamp $t$, the proposed SOFR module $f_{SOFR}$ takes the current frame $x_t$ and the reconstructed reference frame $\hat{x}_{t-1}$ to generate a refined optical flow $v_t$. This flow is then coded by $f_{motion}$ to produce $\hat{v}_t$ for warping the temporal context $C_{t-1}$. The right panel details the internal structure of $f_{SOFR}$. It generates dual-granularity motion candidates Coarse-Grained Flow and Fine-Grained Flow via parallel branches. A Rate-Aware Control module utilizing the Q-Index dynamically guides this generation and their subsequent fusion, ensuring an adaption to varying bitrates.}
  \label{fig:framework}
\end{figure*}

To address these limitations, we propose SOFR, a Scale-driven Online Flow Refinement module for neural video compression. Unlike parameter-updating methods, our approach is entirely \textit{training-free} and serves as a \textit{plug-and-play} module for existing NVCs. The core insight is that motion estimation errors often vary across scales: coarse scales capture global motion trends, while fine scales preserve local details. Therefore, we introduce a dual-granularity fusion mechanism that generates a soft mask to synergize coarse and fine optical flows during inference. Furthermore, to adapt to varying bitrates levels, we design a \textit{rate-aware} strategy. This mechanism dynamically modulates the down-sampling scale of the coarse candidate and the fusion bias based on the quantization parameter, and incorporates a reliability check to filter out ineffective coarse flows. This ensures the preservation of fine-grained details at high bitrates while prioritizing structural information at low bitrates.

The main contributions of this paper are summarized as follows:
\begin{itemize}
    \item We propose a plug-and-play online flow refinement module that effectively bridges the domain gap by fusing multi-granularity motion flows via a soft fusion mask derived from warping accuracy.
    \item We develop a rate-aware strategy that dynamically modulates the down-sampling scale of the coarse candidate and the fusion bias based on the quantization parameter, and incorporates a reliability check to filter out ineffective coarse candidates.
    \item Extensive experiments on USTC-TD dataset verify the generalization of our method to advanced NVCs like DCVC-SDD, DCVC-FM and EHVC, notably achieving average BD-Rate reductions of 2.84\% (PSNR) and 4.05\% (MS-SSIM) on DCVC-FM.
\end{itemize}

\section{Related Work}

\subsection{Neural Video Compression}
In recent years, end-to-end neural video codecs (NVCs) have evolved rapidly, challenging the dominance of traditional standards like H.265/HEVC~\cite{sullivan2012overview} and H.266/VVC~\cite{bross2021overview}. Early works such as DVC~\cite{lu2019dvc} introduced the residual coding paradigm, where optical flow is used for motion compensation and the residuals are compressed by autoencoders. Building upon this, subsequent approaches like DCVC~\cite{li2021deep} and its advanced variants \cite{sheng2022temporal,lin2022dmvc,li2023neural,li2024neural,ye2024deep,sheng2024spatial,jia2025towards,sheng2025bi,zhai2025lbvc,11510469,11510435} have significantly enhanced coding efficiency by incorporating conditional coding mechanisms and exploring more effective predicted features. Specifically, these improvements are generally achieved through more precise feature alignment in motion estimation and compensation, the mining and enhancement of contextual conditional information, and the design of high-efficiency entropy models. Furthermore, recent studies have also investigated diverse architectures, ranging from implicit neural representations~\cite{11462194} that enable fast decoding, to generative models~\cite{SIP-20250056} targeting extremely low bitrates, to further push the boundaries of rate-distortion performance.

\subsection{Optical Flow Optimization in NVC}
Motion estimation is pivotal for video coding performance. To enhance robustness, Agustsson et al.~\cite{agustsson2020scale} introduced a generalized warping operator with a learnable scale to model uncertainty without pre-trained flows. 
Addressing the domain gap via online adaptation, Tang et al.~\cite{tang2024offline} combined offline VVC distillation with online flow latent optimization, while Cong et al.~\cite{cong2025integrating} dynamically adjusted sampling factors to exploit spatio-temporal redundancy. 
To reduce the computational overhead of such updates, Oh et al.~\cite{oh2024parameter} and Chen et al.~\cite{chen2024group} proposed inserting lightweight adapter modules for efficient instance-specific fine-tuning, with the latter employing a group-aware strategy to mitigate error propagation. 
Regarding inference-time optimization for bi-directional coding, Zhai et al.~\cite{zhai2025lbvc} and Yilmaz et al.~\cite{yilmaz2024motion} dynamically selected flow resolutions based on motion complexity. 
Although Bilican et al.~\cite{bilican2025content} explored content-adaptive inference to map motion vectors into the training distribution, their method relies on exhaustively searching for the optimal downsampling factor from a dense set of over 30 candidates (ranging from 1 to 8.75), which leads to excessive computational latency.

\begin{figure*}[t]
  \centering
  \includegraphics[width=0.8\linewidth]{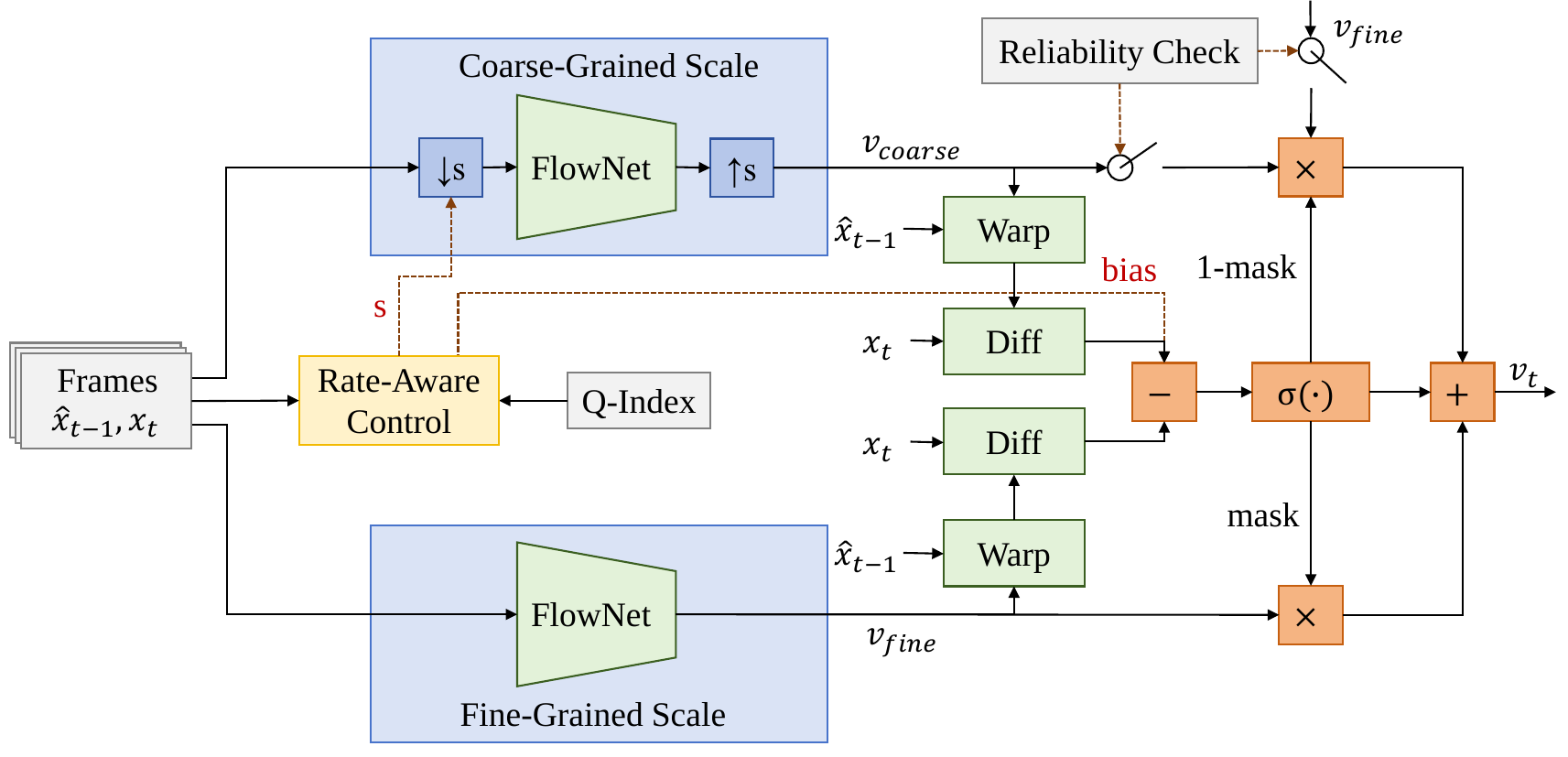} 
  \caption{Illustration of the proposed Scale-Driven Online Flow Refinement (SOFR) module. The architecture employs parallel branches to produce coarse-grained flow $v_{coarse}$ and fine-grained flow $v_{fine}$. This generation is dynamically modulated by a Rate-Aware Control unit, which configures the down-sampling scale $s$ and decision bias according to the Q-Index, and incorporates a reliability check to filter out ineffective coarse candidates. Finally, the candidates are fused into the output $v_t$ via a spatial mask derived from the bias-corrected warping accuracy.}
  \label{fig:sofr_detail}
\end{figure*}

\section{Proposed Method}
\label{sec:method}

\subsection{Overview}
As illustrated in Fig.~\ref{fig:framework}, we integrate our proposed method into a typical contextual Neural Video Codec (NVC). Let $x_t$ denote the current input frame, $\hat{x}_{t-1}$ the reconstructed previous frame, and $C_{t-1}$ the propagated temporal context feature. In the encoding pipeline, the motion estimation module $f_{motion}$ first computes the optical flow $v_t$ between $x_t$ and $\hat{x}_{t-1}$. This motion vector is then compressed and reconstructed into $\hat{v}_t$, which is utilized to warp the temporal feature $F_{t-1}$ to generate current priors $C_{t-1}$ through $f_{context}$ module. Conditioned on this motion-aligned context priors, the frame encoder module $f_{frame}$ takes $x_t$ as input to yield the reconstructed frame $\hat{x}_t$, while simultaneously generating the propagation feature $F_t$ for the next timestamp.

To bridge the domain gap without updating model parameters, we introduce the Scale-Driven Online Flow Refinement (SOFR) module, denoted as $f_{SOFR}$. As shown in the diagram, $f_{SOFR}$ functions as a plug-and-play component for motion estimation during inference. The proposed method consists of two stages. First, the Scale-Driven Dual-Granularity Flow Generation produces coarse-grained and fine-grained motion candidates to serve as complementary motion flows. Subsequently, the Rate-Aware Fusion Strategy dynamically synthesizes the final refined flow field $v_t$ to adapt to varying bitrates. The detailed formulations of these two components are presented in the following subsections.

\subsection{Scale-Driven Dual-Granularity Flow Generation}
The core insight of SOFR is that motion estimation errors show distinct characteristics across spatial scales. While fine-grained flows preserve local structural details, they are vulnerable to large displacements and outliers. Conversely, coarse-grained flows offer robust structural information. To exploit this complementarity, we generate two distinct motion flow candidates in parallel.

\textbf{Fine-Grained Flow.} We first compute optical flow at the original resolution using the pre-trained motion estimation network $F_\theta$. This captures intricate motion details but may fail in regions with complex patterns like large displacements:
\begin{equation}
v_{fine} = F_\theta(x_t, \hat{x}_{t-1})
\end{equation}

\textbf{Coarse-Grained Flow.} To capture dominant motion trends, we introduce a scale-driven down-sampling operation. The down-sampling scale $s$ is dynamically determined by the Rate-Aware Control module. Let $\downarrow_s$ denote the bilinear down-sampling operator and $\uparrow_s$ the corresponding bilinear up-sampling. The coarse flow is computed as:
\begin{equation}
v_{coarse} = \uparrow_s \left( F_\theta(\downarrow_s(x_t), \downarrow_s(\hat{x}_{t-1})) \right) \times s
\end{equation}
where the multiplication by $s$ rescales the magnitude of the motion vectors to the original domain. This process effectively filters out high-frequency noise and provides a smoother motion flow that is easier to compression.

\begin{table*}[t]
\caption{BD-Rate performance comparison on the USTC-TD dataset. Models integrated with our SOFR and optimized for PSNR are marked with an asterisk ($*$), while those optimized for MS-SSIM are marked with a dagger ($^\dagger$). Negative values indicate bitrate savings compared to their respective unoptimized baselines.}
\label{tab:ustc_results}
\centering
\renewcommand{\arraystretch}{1.2}
\setlength{\tabcolsep}{0.6mm}

\resizebox{\linewidth}{!}{
\begin{tabular}{ccccccccccccc} 
\toprule
\textbf{Method} & \textbf{Badminton} & \textbf{BasketDrill} & \textbf{BasketPass} & \textbf{BicycleDriving} & \textbf{Dancing} & \textbf{FourPeople} & \textbf{ParkWalking} & \textbf{Running} & \textbf{ShakeHands} & \textbf{Snooker} & \textbf{Average} & \textbf{Coding Time} \\
\midrule

DCVC-FM$^*$        & \underline{-0.88} & \underline{-9.65} & \underline{-6.81} & \underline{-2.06} & 0.18 & 0.19 & \underline{-4.40} & 0.04 & \underline{-2.68} & \underline{-2.34} & \underline{\textbf{-2.84}} & \multirow{2}{*}{1.0240$\times$} \\
DCVC-FM$^\dagger$  & \underline{-0.56} & \underline{-13.31} & \underline{-8.53} & \underline{-3.07} & 0.08 & \underline{-0.18} & \underline{-7.43} & 0.24 & \underline{-4.37} & \underline{-3.34} & \textbf{\underline{-4.05}} & \\
DCVC-SDD$^*$       & 0.14 & \underline{-0.87} & \underline{-0.27} & \underline{-6.61} & 0.03 & \underline{-0.01} & \underline{-0.32} & 0.09 & 0.27 & \underline{-2.61} & \underline{\textbf{-0.97}} & \multirow{2}{*}{1.0144$\times$} \\
DCVC-SDD$^\dagger$ & 0.01 & \underline{-0.81} & 0.00 & \underline{-3.74} & 0.01 & \underline{-0.03} & 0.08 & \underline{-0.06} & 0.33 & \underline{-0.15} & \textbf{\underline{-0.44}} & \\
EHVC$^*$           & 0.22 & \underline{-3.78} & \underline{-1.78} & \underline{-3.59} & \underline{-0.06} & \underline{-0.04} & 2.27 & 0.01 & \underline{-0.26} & \underline{-2.11} & \underline{\textbf{-0.91}} & \multirow{2}{*}{1.0181$\times$} \\
EHVC$^\dagger$     & 0.02 & \underline{-4.67} & \underline{-2.32} & \underline{-4.68} & 0.05 & \underline{-0.03} & 1.89 & \underline{-0.03} & \underline{-0.61} & \underline{-2.30} & \textbf{\underline{-1.27}} & \\

\bottomrule 
\end{tabular}
}
\end{table*}

\begin{figure*}[t]
  \centering
  \includegraphics[width=0.95\linewidth]{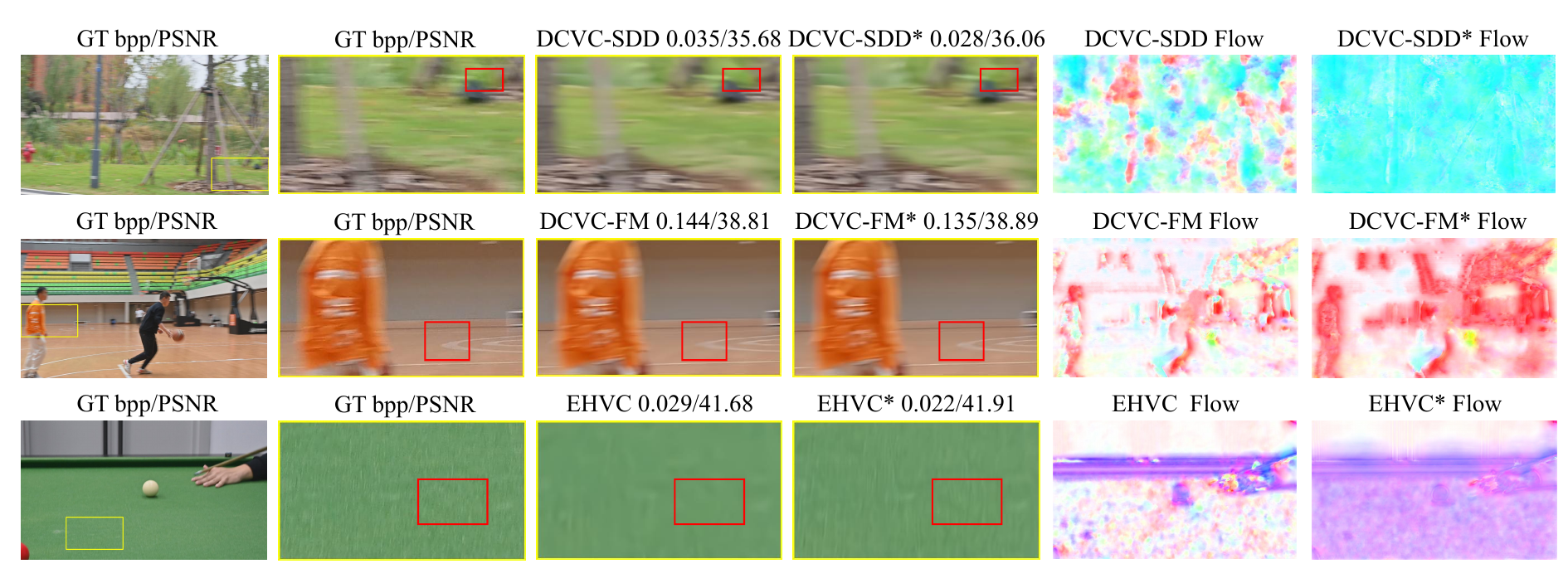} 
  \caption{Visual comparison on the USTC-TD dataset, specifically selecting the 69th frame of the \textit{BicycleDriving} sequence for DCVC-SDD, the 23rd frame of the \textit{BasketDrill} sequence for DCVC-FM, and the 15th frame of the \textit{Snooker} sequence for EHVC. The columns from left to right display the ground truth, the reconstruction details comparison, and the optical flow comparison. As observed in the flow maps, the baseline suffers from estimation noise in dynamic regions, whereas SOFR produces precise and coherent motion flows. This accurate motion propagation enables our method to achieve higher reconstruction quality at a lower bitrate.}
  \label{fig:visual_comp}
\end{figure*}

\subsection{Rate-Aware Fusion Strategy}
To effectively synthesize the dual-granularity flows, we propose a Rate-Aware Fusion Strategy that adapts to the varying bitrates. As illustrated in Fig.~\ref{fig:sofr_detail}, this module takes the quantization parameter (Q-Index) as input to control the generation and fusion process via two hyper-parameters: the down-sampling scale $s$ and the decision bias $b$.

\textbf{Rate-Aware Control.} We categorize the coding scenarios into high bitrates and low bitrates states based on the quality levels defined by the Q-Index.

1) \textit{High Bitrates Mode:} This corresponds to the upper half of the quality levels. In this regime, we prioritize fine-grained detail preservation. We set the scale to $s=2$ to retain more spatial information and assign a lower bias $b=0.005$ to encourage the usage of fine-grained details.

2) \textit{Low Bitrates Mode:} This corresponds to the lower half of the quality levels. In this regime, minimizing the coding cost of motion vectors is critical. We set the scale to $s=4$ to strongly filter noise and generate a compact motion field. Simultaneously, we increase the bias to $b=0.03$. By adding this larger penalty to the fine flow error, we guide the network to prefer the smoother coarse flow, thereby saving bits.

\textbf{Reliability Check and Fusion.}
Subsequently, to synthesize the final motion flow, we evaluate the pixel-wise prediction consistency. Let $\mathcal{W}(\cdot, v)$ denote the warping operation. We first compute the prediction error maps $E_{k}$ for both fine and coarse candidates ($k \in \{fine, coarse\}$):
\begin{equation}
E_{k} = | x_t - \mathcal{W}(\hat{x}_{t-1}, v_{k}) |
\end{equation}

To avoid performance degradation caused by dual-granularity flow fusion on sequences where additional adjustment is unnecessary, we employ a global reliability check. We calculate the Mean Absolute Difference (MAD) for each candidate as $\bar{E}_{k} = \text{mean}(E_{k})$. The effective coarse candidate $\tilde{v}_{coarse}$ is determined by comparing these global metrics:
\begin{equation}
\tilde{v}_{coarse} = 
\begin{cases} 
v_{fine}, & \text{if } \bar{E}_{coarse} \ge \bar{E}_{fine} \\
v_{coarse}, & \text{otherwise}
\end{cases}
\end{equation}

This ensures that if the coarse flow yields larger errors, the system effectively reverts to the fine-grained flow. Based on the validated candidates, we derive a soft fusion mask $M$ incorporating the rate-aware bias $b$:
\begin{equation}
M = \sigma \left( \alpha \cdot (\tilde{E}_{coarse} - E_{fine} - b) \right)
\end{equation}
where $\tilde{E}_{coarse}$ corresponds to the error of $\tilde{v}_{coarse}$, and $\alpha$ is a sharpness factor. Finally, the refined motion field $v_t$ is synthesized as:
\begin{equation}
v_t = M \odot v_{fine} + (1 - M) \odot \tilde{v}_{coarse}
\end{equation}

\begin{table*}[t]
\caption{Ablation results in terms of PSNR on the first 96 frames of the USTC-TD dataset relative to the DCVC-FM baseline under Intra Period -1 settings. The study analyzes the impact of fixed versus dynamic scale strategies and the effectiveness of the Rate-Aware Fusion mechanism.}
\label{tab:ablation}
\centering
\renewcommand{\arraystretch}{1.2} 
\setlength{\tabcolsep}{1.5mm}      

\begin{tabular}{ccccccc}
\toprule
\textbf{Method} & \textbf{Model} & \textbf{Coarse Scale ($s$)} & \textbf{Rate-Aware Control} & \textbf{Reliability Check} & \textbf{BD-Rate (\%)} & \textbf{Coding Time} \\
\midrule
Baseline & Anchor & -- & -- & -- & 0.00 & 1.0000$\times$ \\
\midrule

\multirow{3}{*}{Dual-Granularity Flow} & $M_a$ & 2 & No & No & \underline{-2.15} & -- \\
                                    & $M_b$ & 4 & No & No & 3.01 & -- \\
                                    & $M_c$ & 8 & No & No & 30.04 & -- \\
\midrule

\multirow{2}{*}{Rate-Aware Fusion} 
                                      & $M_d$ & Dynamic & Yes & No & \underline{-2.19} & -- \\
                                      & \textbf{$M_e$} & \textbf{Dynamic} & \textbf{Yes} & \textbf{Yes} & \underline{\textbf{-2.84}} & \textbf{1.0240$\times$} \\

\bottomrule
\end{tabular}
\end{table*}

\section{Experiments}

\subsection{Experimental Settings}
\textbf{Test Datasets.} To validate the effectiveness of the proposed SOFR module, we conduct experiments on USTC-TD datasets~\cite{li2025ustc}. Recognized as a high quality benchmark from the 2020s, USTC-TD offers a comprehensive representation of modern video content characteristics. It features 10 sequences with diverse motion patterns and textures, all at a resolution of $1920 \times 1080$. All raw YUV sequences are converted to the RGB color space using the BT.709 standard. During inference, we apply padding to the input frames to align with the network's down-sampling requirements.

\textbf{Implementation Details and Baselines.} Since SOFR functions as a training-free inference optimization strategy, we directly integrate it into three representative SOTA baselines from recent years: DCVC-SDD~\cite{sheng2024spatial}, DCVC-FM~\cite{li2024neural}, and EHVC~\cite{liao2025ehvc}. For evaluation, we strictly adhere to the optimal configurations provided in their official implementations. We set the Intra Period (IP) to 32 for DCVC-SDD, while configuring DCVC-FM and EHVC with an IP of -1. We conduct all experiments on the first 96 frames of each sequence and perform actual encoding and decoding processes to generate real binary bitstreams for precise bitrate calculation.

\textbf{Evaluation Metrics.} We measure the bitrate using bits-per-pixel (bpp). The reconstruction quality is assessed by both PSNR and MS-SSIM. To quantify the coding efficiency improvement, we utilize the Bjontegaard Delta Rate (BD-Rate)~\cite{bjontegaard2001calculation}, where negative values indicate bitrate savings at the same quality level.

\subsection{Main Results}

We evaluate the R-D performance of SOFR integrated with the aforementioned baselines. Table~\ref{tab:ustc_results} summarizes the BD-Rate savings on the USTC-TD dataset. As shown, SOFR consistently boosts the compression efficiency across all three SOTA baselines in terms of both PSNR and MS-SSIM. In detail, it brings average bitrate savings of 0.97\% (PSNR) and 0.44\% (MS-SSIM) to DCVC-SDD, and reduces the BD-Rate of EHVC by 0.91\% (PSNR) and 1.27\% (MS-SSIM). Most notably, when integrated into DCVC-FM, our method achieves substantial average gains of 2.84\% and 4.05\% in terms of PSNR and MS-SSIM, respectively. Specifically, complex sequences like \textit{BasketDrill} and \textit{BasketPass} show remarkable savings of up to 9.65\% (PSNR) / 13.31\% (MS-SSIM) and 6.81\% (PSNR) / 8.53\% (MS-SSIM) with negligible computational overhead. These extensive results across diverse baselines demonstrate the wide applicability and robustness of the proposed strategy.

\subsection{Ablation Studies}
To strictly isolate and verify the contribution of each component in SOFR, we conduct ablation experiments on the USTC-TD dataset using DCVC-FM as the anchor, evaluated on the first 96 frames with an Intra Period of -1. The quantitative results of five variants, measured in terms of BD-Rate (PSNR), are summarized in Table \ref{tab:ablation}.

\textbf{Impact of Dual-Granularity Flow.} We first investigate the Dual-Granularity Flow with fixed down-sampling scales. As shown in the first group, Model $M_a$ ($s=2$) achieves a bitrate saving of 2.15\%. However, simply increasing the down-sampling ratio proves detrimental: $M_b$ ($s=4$) and $M_c$ ($s=8$) result in performance degradation with BD-Rate increases of 3.01\% and 30.04\%, respectively. This indicates that while coarse flows can be beneficial, excessive down-sampling indiscriminately destroys motion structure in complex regions, leading to inaccurate flow estimation.

\textbf{Effectiveness of Rate-Aware Fusion.} We then evaluate the proposed Rate-Aware strategies. 
Model $M_d$, which introduces the Rate-Aware Control like dynamic scale $s$ and bias $b$ based on Q-Index, achieves a saving of 2.19\%, outperforming the best fixed-scale variant $M_a$. This validates the benefit of adaptively adjusting the fusion strategy according to bitrates.
Crucially, equipping the system with the Reliability Check $M_e$ further boosts the performance to 2.84\%. Comparing $M_e$ with $M_d$, the gating mechanism provides a substantial gain, confirming its ability to identify and filter out low-quality coarse flows, which caused the losses in $M_b$.

\subsection{Case Analysis} 
We select representative frames from the USTC-TD dataset to provide a qualitative comparison in Fig.~\ref{fig:visual_comp}. Specifically, we select the 69th frame of the \textit{BicycleDriving} sequence for DCVC-SDD, the 23rd frame of the \textit{BasketDrill} sequence for DCVC-FM, and the 15th frame of the \textit{Snooker} sequence for EHVC. As illustrated, our proposed SOFR achieves higher reconstruction quality while consuming a significantly lower bpp compared to the baselines. Looking specifically at the optical flow visualization, the baseline model tends to produce noisy estimations in regions with complex dynamics. In contrast, SOFR effectively optimizes motion details by dynamically fusing dual-granularity flows, yielding a more precise and coherent motion flow. This improvement directly translates to the pixel domain, where the reconstructed frame shows sharper structural edges and fewer artifacts in the highlighted areas.

\subsection{Complexity Analysis}
Computational efficiency is critical for online optimization. We evaluate the complexity on an NVIDIA A100 GPU (40G). As reported in Table~\ref{tab:ablation}, SOFR increases the entire coding time from $1.0000\times$ to $1.0240\times$, representing a negligible overhead, while the decoding time remains unchanged. In contrast to time-consuming gradient-based iterative methods, our feed-forward design ensures high efficiency for practical deployment.

\section{Conclusion}
\label{sec:conclusion}

In this paper, we proposed SOFR, a training-free module designed to bridge the motion domain gap in neural video codecs. By integrating coarse-grained structural stability with fine-grained local details via a warping-accuracy based soft mask, our method effectively rectifies motion estimation errors. Furthermore, the proposed rate-aware strategy ensures robust performance across varying bitrates by dynamically adjusting fusion parameters and filtering ineffective candidates through a reliability check. Extensive experiments verify that SOFR serves as an efficient plug-and-play solution, consistently improving the rate-distortion performance of state-of-the-art frameworks, including DCVC-SDD, DCVC-FM, and EHVC, with negligible computational overhead.

\bibliographystyle{IEEEbib}
\bibliography{icme2026references}

@article{bross2021overview,
  title={Overview of the versatile video coding (VVC) standard and its applications},
  author={Bross, Benjamin and Wang, Ye-Kui and Ye, Yan and Liu, Shan and Chen, Jianle and Sullivan, Gary J and Ohm, Jens-Rainer},
  journal={IEEE Transactions on Circuits and Systems for Video Technology},
  volume={31},
  number={10},
  pages={3736--3764},
  year={2021},
  publisher={IEEE}
}

@inproceedings{tang2024offline,
  title={Offline and online optical flow enhancement for deep video compression},
  author={Tang, Chuanbo and Sheng, Xihua and Li, Zhuoyuan and Zhang, Haotian and Li, Li and Liu, Dong},
  booktitle={Proceedings of the AAAI Conference on Artificial Intelligence},
  volume={38},
  number={6},
  pages={5118--5126},
  year={2024}
}

@inproceedings{oh2024parameter,
  title={Parameter-efficient instance-adaptive neural video compression},
  author={Oh, Seungjun and Yang, Hyunmo and Park, Eunbyung},
  booktitle={Proceedings of the Asian Conference on Computer Vision},
  pages={250--267},
  year={2024}
}

@inproceedings{chen2024group,
  title={Group-aware parameter-efficient updating for content-adaptive neural video compression},
  author={Chen, Zhenghao and Zhou, Luping and Hu, Zhihao and Xu, Dong},
  booktitle={Proceedings of the 32nd ACM International Conference on Multimedia},
  pages={11022--11031},
  year={2024}
}

@inproceedings{cong2025integrating,
  title={Integrating Adaptive Sampling for Optimal Learned Video Compression},
  author={Cong, Wuyang and Kong, Yuzhuo and Lu, Ming and Wang, Lizhong and Shi, Weijing and Ma, Zhan},
  booktitle={IEEE International Conference on Acoustics, Speech and Signal Processing},
  year={2025}
}

@inproceedings{zhai2025lbvc,
  title={L-LBVC: Long-Term Motion Estimation and Prediction for Learned Bi-Directional Video Compression},
  author={Zhai, Yongqi and Tang, Luyang and Jiang, Wei and Yang, Jiayu and Wang, Ronggang},
  booktitle={IEEE Data Compression Conference},
  pages={53--62},
  year={2025}
}

@article{bilican2025content,
  title={Content-Adaptive Inference for State-of-the-art Learned Video Compression},
  author={Bilican, Ahmet and Y{\i}lmaz, M Ak{\i}n and Tekalp, A Murat},
  journal={IEEE Open Journal of Signal Processing},
  year={2025},
  publisher={IEEE}
}

@inproceedings{yilmaz2024motion,
  title={Motion-adaptive inference for flexible learned b-frame compression},
  author={Yilmaz, M Akin and Ulas, O Ugur and Bilican, Ahmet and Tekalp, A Murat},
  booktitle={IEEE International Conference on Image Processing},
  pages={1760--1766},
  year={2024}
}

@inproceedings{agustsson2020scale,
  title={Scale-space flow for end-to-end optimized video compression},
  author={Agustsson, Eirikur and Minnen, David and Johnston, Nick and Balle, Johannes and Hwang, Sung Jin and Toderici, George},
  booktitle={Proceedings of the IEEE/CVF Conference on Computer Vision and Pattern Recognition},
  pages={8503--8512},
  year={2020}
}

@article{sullivan2012overview,
  title={Overview of the high efficiency video coding (HEVC) standard},
  author={Sullivan, Gary J and Ohm, Jens-Rainer and Han, Woo-Jin and Wiegand, Thomas},
  journal={IEEE Transactions on Circuits and Systems for Video Technology},
  volume={22},
  number={12},
  pages={1649--1668},
  year={2012},
  publisher={IEEE}
}

@article{li2021deep,
  title={Deep contextual video compression},
  author={Li, Jiahao and Li, Bin and Lu, Yan},
  journal={Advances in Neural Information Processing Systems},
  volume={34},
  pages={18114--18125},
  year={2021}
}

@inproceedings{li2023neural,
  title={Neural video compression with diverse contexts},
  author={Li, Jiahao and Li, Bin and Lu, Yan},
  booktitle={Proceedings of the IEEE/CVF conference on computer vision and pattern recognition},
  pages={22616--22626},
  year={2023}
}

@article{sheng2022temporal,
  title={Temporal context mining for learned video compression},
  author={Sheng, Xihua and Li, Jiahao and Li, Bin and Li, Li and Liu, Dong and Lu, Yan},
  journal={IEEE Transactions on Multimedia},
  volume={25},
  pages={7311--7322},
  year={2022},
  publisher={IEEE}
}

@inproceedings{hu2022coarse,
  title={Coarse-to-fine deep video coding with hyperprior-guided mode prediction},
  author={Hu, Zhihao and Lu, Guo and Guo, Jinyang and Liu, Shan and Jiang, Wei and Xu, Dong},
  booktitle={Proceedings of the IEEE/CVF Conference on Computer Vision and Pattern Recognition},
  pages={5921--5930},
  year={2022}
}

@inproceedings{li2024neural,
  title={Neural video compression with feature modulation},
  author={Li, Jiahao and Li, Bin and Lu, Yan},
  booktitle={Proceedings of the IEEE/CVF Conference on Computer Vision and Pattern Recognition},
  pages={26099--26108},
  year={2024}
}

@inproceedings{ye2024deep,
  title={Deep video compression with scaled hierarchical bi-directional motion model},
  author={Ye, Feng and Zhang, Li and Jia, Chuanmin},
  booktitle={Proceedings of the 32nd ACM International Conference on Multimedia},
  pages={11244--11247},
  year={2024}
}

@inproceedings{jia2025emerging,
  title={Emerging advances in learned video compression: models, systems and beyond},
  author={Jia, Chuanmin and Ye, Feng and Ma, Siwei and Gao, Wen and Sun, Huifang and Chiariglione, Leonardo},
  booktitle={Proceedings of the Thirty-Fourth International Joint Conference on Artificial Intelligence},
  pages={10490--10498},
  year={2025}
}

@article{sheng2024spatial,
  title={Spatial decomposition and temporal fusion based inter prediction for learned video compression},
  author={Sheng, Xihua and Li, Li and Liu, Dong and Li, Houqiang},
  journal={IEEE Transactions on Circuits and Systems for Video Technology},
  volume={34},
  number={7},
  pages={6460--6473},
  year={2024},
  publisher={IEEE}
}

@inproceedings{jia2025towards,
  title={Towards practical real-time neural video compression},
  author={Jia, Zhaoyang and Li, Bin and Li, Jiahao and Xie, Wenxuan and Qi, Linfeng and Li, Houqiang and Lu, Yan},
  booktitle={Proceedings of the Computer Vision and Pattern Recognition Conference},
  pages={12543--12552},
  year={2025}
}

@article{sheng2025bi,
  title={Bi-directional deep contextual video compression},
  author={Sheng, Xihua and Li, Li and Liu, Dong and Wang, Shiqi},
  journal={IEEE Transactions on Multimedia},
  year={2025},
  publisher={IEEE}
}

@INPROCEEDINGS{11510435,
  author={Zhang, Tiange and Huang, Zhimeng and Meng, Xiandong and Zhang, Kai and Deng, Zhipin and Ma, Siwei},
  booktitle={IEEE Data Compression Conference}, 
  title={L-STEC: Learned Video Compression with Long-Term Spatio-Temporal Enhanced Context}, 
  year={2026},
  volume={},
  number={},
  pages={13-22},
  doi={10.1109/DCC66757.2026.00009}}

@INPROCEEDINGS{11510469,
  author={Zhang, Tiange and Meng, Xiandong and Ma, Siwei},
  booktitle={IEEE Data Compression Conference}, 
  title={Content Adaptive Based Motion Alignment Framework for Learned Video Compression}, 
  year={2026},
  volume={},
  number={},
  pages={486-486},
  doi={10.1109/DCC66757.2026.00113}}

@INPROCEEDINGS{11561917,
  author={Zhang, Tiange and Lin, Rongqun and Meng, Xiandong and Wang, Haofeng and Tian, Xing and Zhang, Qi and Ma, Siwei},
  booktitle={IEEE International Symposium on Circuits and Systems}, 
  title={Neural Video Compression with Domain Transfer}, 
  year={2026},
  volume={},
  number={},
  pages={1551-1555},
  doi={10.1109/ISCAS66217.2026.11561917}}

@INPROCEEDINGS{11462194,
  author={Zhang, Tiange and Wang, Haofeng and Meng, Xiandong and Zhang, Kai and Deng, Xuan and Huang, Zhimeng and Ma, Siwei},
  booktitle={IEEE International Conference on Acoustics, Speech and Signal Processing}, 
  title={Chnerv: Condition Enhanced Hybrid Neural Representation for Videos}, 
  year={2026},
  volume={},
  number={},
  pages={8377-8381},
  doi={10.1109/ICASSP55912.2026.11462194}}

@article{SIP-20250056,
	url = {http://dx.doi.org/10.1561/116.20250056},
	year = {2025},
	volume = {14},
	journal = {APSIPA Transactions on Signal and Information Processing},
	title = {Generative Coding: Promise and Challenges},
	doi = {10.1561/116.20250056},
	author = {Siwei Ma and Shenpeng Song and Bolin Chen and Qi Mao and Xiaohan Fang and Chuanmin Jia and Shiqi Wang}
}

@inproceedings{lu2019dvc,
  title={Dvc: An end-to-end deep video compression framework},
  author={Lu, Guo and Ouyang, Wanli and Xu, Dong and Zhang, Xiaoyun and Cai, Chunlei and Gao, Zhiyong},
  booktitle={Proceedings of the IEEE/CVF Conference on Computer Vision and Pattern Recognition},
  pages={11006--11015},
  year={2019}
}

@article{lin2022dmvc,
  title={DMVC: Decomposed motion modeling for learned video compression},
  author={Lin, Kai and Jia, Chuanmin and Zhang, Xinfeng and Wang, Shanshe and Ma, Siwei and Gao, Wen},
  journal={IEEE Transactions on Circuits and Systems for Video Technology},
  volume={33},
  number={7},
  pages={3502--3515},
  year={2022},
  publisher={IEEE}
}

@article{ma2019image,
  title={Image and video compression with neural networks: A review},
  author={Ma, Siwei and Zhang, Xinfeng and Jia, Chuanmin and Zhao, Zhenghui and Wang, Shiqi and Wang, Shanshe},
  journal={IEEE Transactions on Circuits and Systems for Video Technology},
  volume={30},
  number={6},
  pages={1683--1698},
  year={2019},
  publisher={IEEE}
}

@article{li2025ustc,
  title={Ustc-td: A test dataset and benchmark for image and video coding in 2020s},
  author={Li, Zhuoyuan and Liao, Junqi and Tang, Chuanbo and Zhang, Haotian and Li, Yuqi and Bian, Yifan and Sheng, Xihua and Feng, Xinmin and Li, Yao and Gao, Changsheng and others},
  journal={IEEE Transactions on Multimedia},
  year={2025},
  publisher={IEEE}
}

@article{bjontegaard2001calculation,
  title={Calculation of average PSNR differences between RD-curves},
  author={Bjontegaard, Gisle},
  journal={ITU-T SG16, Doc. VCEG-M33},
  year={2001}
}

@inproceedings{liao2025ehvc,
  title={EHVC: Efficient Hierarchical Reference and Quality Structure for Neural Video Coding},
  author={Liao, Junqi and Wu, Yaojun and Lin, Chaoyi and Deng, Zhipin and Li, Li and Liu, Dong and Sun, Xiaoyan},
  booktitle={Proceedings of the 33rd ACM International Conference on Multimedia},
  pages={12083--12091},
  year={2025}
}

@article{LIN2023126396,
title = {Multiple hypotheses based motion compensation for learned video compression},
journal = {Neurocomputing},
volume = {548},
pages = {126396},
year = {2023},
issn = {0925-2312},
doi = {https://doi.org/10.1016/j.neucom.2023.126396},
url = {https://www.sciencedirect.com/science/article/pii/S0925231223005192},
author = {Rongqun Lin and Meng Wang and Pingping Zhang and Shiqi Wang and Sam Kwong}
}

@article{xue2019video,
  title={Video enhancement with task-oriented flow},
  author={Xue, Tianfan and Chen, Baian and Wu, Jiajun and Wei, Donglai and Freeman, William T},
  journal={International Journal of Computer Vision},
  volume={127},
  number={8},
  pages={1106--1125},
  year={2019},
  publisher={Springer}
}

\end{document}